\documentclass[letterpaper, 10 pt, conference]{ieeeconf}  

\IEEEoverridecommandlockouts                              

\usepackage{graphics} 
\usepackage{epsfig} 
\usepackage{mathptmx} 
\usepackage{times} 
\usepackage{amsmath} 
\usepackage{amssymb}  
\usepackage{algorithm}
\usepackage{algpseudocode}
\usepackage{xcolor}

\title{\LARGE \bf
Deep Object-Centric Policies for Autonomous Driving
}

\author{Dequan Wang$^{1}$, Coline Devin$^{1}$, Qi-Zhi Cai$^{2*}$, Fisher Yu$^{1}$, Trevor Darrell$^{1}$%
\thanks{$^{1}$University of California, Berkeley, $^{2}$Nanjing University}%
\thanks{$^*$Work done while at University of California, Berkeley}
}

\begin{document}

\newcommand{\ra}[1]{\renewcommand{\arraystretch}{#1}}
\newcommand{\dq}[1]{\textcolor{orange}{[DEQUAN: #1 ]}}
\newcommand{\cd}[1]{\textcolor{cyan}{[COLINE: #1 ]}}
\newcommand{\td}[1]{\textcolor{magenta}{[TREVOR: #1 ]}}
\newcommand{\TBD}[1]{\textcolor{red}{[TBD: #1 ]}}

\newcommand{\etal}{\textit{et al}. }
\newcommand{\ie}{\textit{i}.\textit{e}., }
\newcommand{\eg}{\textit{e}.\textit{g}. }

\maketitle
\thispagestyle{empty}
\pagestyle{empty}

\begin{abstract}
While learning visuomotor skills in an end-to-end manner is appealing, deep neural networks are often uninterpretable and fail in surprising ways.
For robotics tasks, such as autonomous driving, models that explicitly represent objects may be more robust to new scenes and provide intuitive visualizations.
We describe a taxonomy of ``object-centric" models which leverage both object instances and end-to-end learning. In the Grand Theft Auto V simulator, we show that object-centric models outperform object-agnostic methods in scenes with other vehicles and pedestrians, even with an imperfect detector. We also demonstrate that our architectures perform well on real-world environments by evaluating on the Berkeley DeepDrive Video dataset, where an object-centric model outperforms object-agnostic models in the low-data regimes.
\end{abstract}

\section{INTRODUCTION}

End-to-end approaches to visuomotor learning are appealing in their ability to discover which features of an observed environment are most relevant for a task,
and to be able to exploit large amounts of training data to discover both a policy and a co-dependent visual representation.
Yet, the key benefit of such approaches---that they learn from task experience---is also their Achilles' heel when it comes to many real-world settings,
where behavioral training data is not unlimited and correct perception of the many rare events that can be encountered is critical for robust performance.

Learning all visual parameters of a visuomotor policy from task reward (or demonstration cloning) places an undue burden on task-level supervision or reward.
In autonomous driving scenarios, for example, an agent should ideally be able to perceive objects and vehicles with a wide range of appearance, even those that are not well represented in a behavioral training set.
Indeed, for many visuomotor tasks, there exist related datasets with supervision for perception tasks, such as detection or segmentation, that do not provide supervision for behaviour learning.
Learning the entire range of vehicle appearance from steering supervision alone, while optimal in the limit of infinite training data, clearly misses the mark in many practical settings. We propose to leverage such datasets when learning driving policies.

Classic approaches to robotic perception have employed separate object detectors to provide a fixed state representation to a rule-based policy.
Multistage methods, such as those which first segment a scene~\cite{muller2018driving},
can avoid some  aspects of the domain transfer problem,
but do not encode discrete objects and thus are limited to holistic reasoning. End-to-end learning with pixel-wise attention can localize specific objects and provide interpretability, but throws away the existence of instances.

We propose an object-centric perception approach to deep control problems, and focus our experimentation on autonomous driving.
Existing end-to-end models are holistic in nature; our approach augments policy learning with explicit representations that provide object-level attention.

In this work, we explore a taxonomy of representations that consider different levels of objects-centric representations, with various discreteness and sparsity.
We define a family of approaches to object-centric models, and provide a comparative evaluation of the benefit of incorporating object knowledge either at a pixel or box level, with either sparse or dense coverage, and with either pooled or concatenated features.

\begin{figure}[t!]
    \begin{center}
        \begin{tabular}{c | c}
            \includegraphics[width=0.45\linewidth]{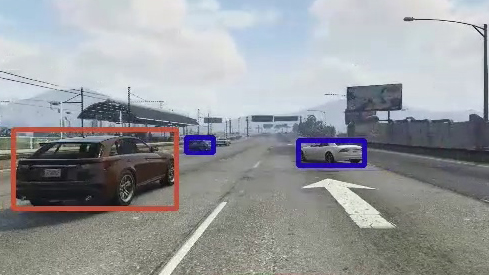} &
            \includegraphics[width=0.45\linewidth]{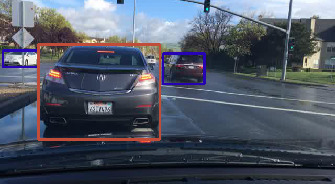} \\
            \includegraphics[width=0.45\linewidth]{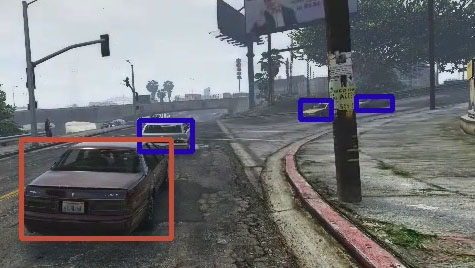} &
            \includegraphics[width=0.45\linewidth]{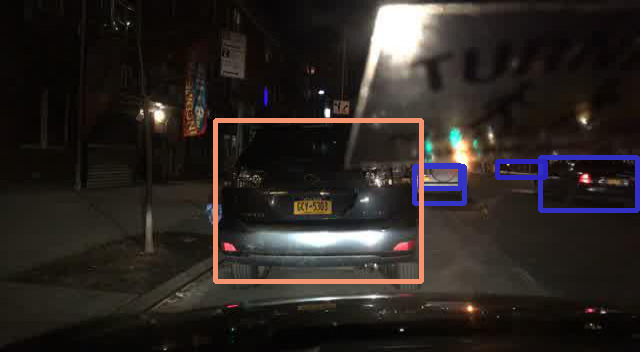} \\
        \end{tabular}
\end{center}
    \caption{Our method uses discrete objects as part of the policy model for driving in traffic. The learned selector identifies the objects most relevant to the policy, which is often the nearest car. We evaluate on both simulated (left, GTAV) and real (right, BDDV) datasets.}
    \label{fig:challenge}
\end{figure}

We evaluate our models in a challenging simulated driving environment with many cars and pedestrians, as well as on real dash-cam data, as shown in Figure~\ref{fig:challenge}.
We show that using a sparse and discrete object-centric representation with a learned per-object attention outperforms previous methods in on-policy evaluations, such models provide the additional interpretability about which objects were determined most relevant to the policy.

\begin{figure*}[ht]
    \begin{center}
        \includegraphics[width=0.8\linewidth]{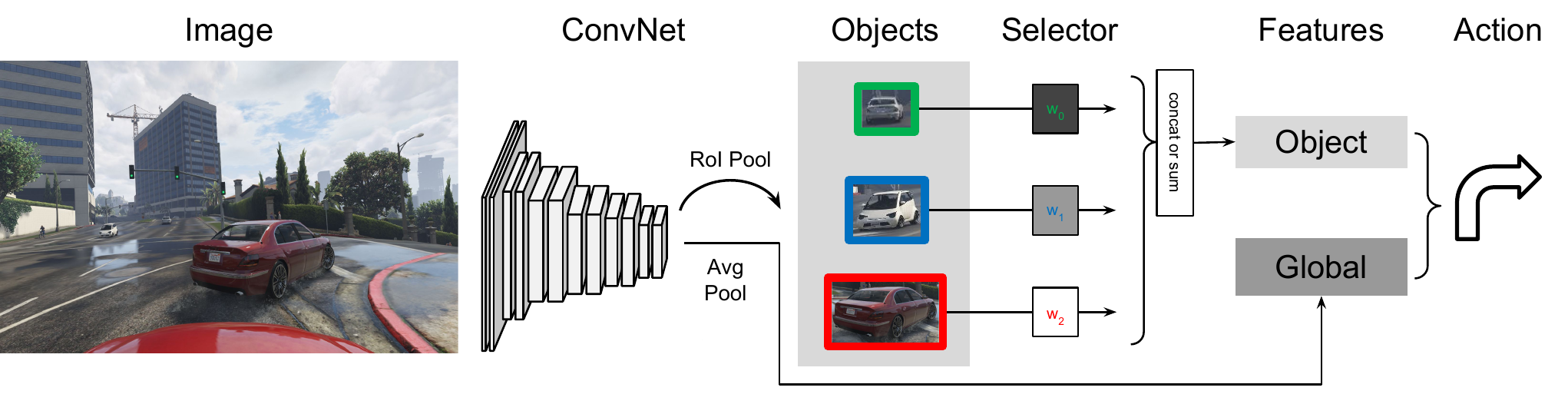}
    \end{center}
    \vspace{-5mm}
    \caption{Overview of our object-centric architecture. An image is first passed through a convolutional network, which outputs RoI pooled features for each object along with globally pooled features for the whole image. Then object-level attention layer calculates the task-oriented importance score for each RoI. The linear policy layer takes both global and object features and predicts action for next step.}
    \label{fig:pipeline}
\end{figure*}

\begin{figure*}[ht]
    \begin{center}
        \includegraphics[width=0.8\linewidth]{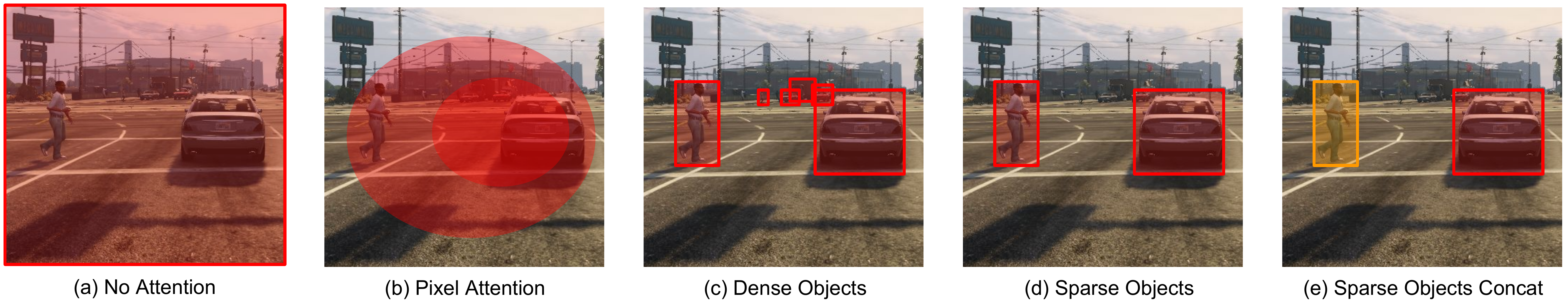}
    \end{center}
    \vspace{-3mm}
    \caption{An illustration of the representation taxonomy we describe in Section~\ref{sec:taxonomy}. (a) shows a global image representation that does not leverage objects. (b) is a continuous (pixel-level) attention that selects salient parts of the image. (c) is a dense and discrete object representation that selects all objects in the scene. (d) is a discrete but sparse object presentation that only selects the objects important for the task, and (e) is a sparse representation that treats each object individually by concatenating instead of averaging the object features.}
    \label{fig:taxonomy}
\end{figure*}

\section{RELATED WORK}

Approaches to robot skill learning face bias/variance trade-offs,
including in the definition of a policy model.
One extreme of this trade-off is to make no assumptions about the structure of the observations,
such as end-to-end behavior cloning from raw sensory data~\cite{bojarski2016end, bojarski2017explaining, xu2017end}.
At the opposite end, one can design a policy structure that is very specific to a particular task,
\eg for driving by calculating margins between cars, encoding lane following, and tracking pedestrians~\cite{huval2015empirical}.
These modular pipelines with rule-based system dominate autonomous driving industry~\cite{thrun2006stanley, urmson2008autonomous, ziegler2014making}.

The first attempt at training an end-to-end driving policy from raw inputs traces back to 1980s with ALVINN~\cite{pomerleau1989alvinn}.
Muller \etal revisited this idea to help off-road mobile robots with obstacle avoidance system~\cite{muller2006off}.
Recently, Bojarski \etal demonstrate the appeal of foregoing structure by training a more advanced convolutional network to imitate demonstrated driving~\cite{bojarski2016end, bojarski2017explaining}.
Xu \etal advocate learning a driving policy from a uncalibrated crowd-sourced video dataset~\cite{xu2017end} and show their model can predict the true actions taken by the drivers from RGB inputs.
Codevilla \etal leverage the idea of conditional imitation learning on high-level command input in order to resolve the ambiguity in action space~\cite{codevilla2017end}.
These end-to-end models, which automatically discover and construct the mapping from sensory input to control output,
reduce the burden of hand-crafting rules and features.
However, these approaches have not yet been shown to work in complex environments, such as intersections with other drivers and pedestrians.

We address how to best represent images for
robotics tasks such as driving.
Muller \etal train a policy model from the semantic segmentation of images, which increases generalization from synthetic to real-world~\cite{muller2018driving}.
Chen \etal provide an additional intermediate stage for end-to-end learning,
which learns the policy on the top of some ConvNet-based measurements,
such as affordance of road/traffic state for driving~\cite{chen2015deepdriving}.
Sauer \etal combine the advantages of conditional learning and affordance~\cite{sauer2018conditional}.
The policy module is built on a set of low-dimensional affordance measurements, with the given navigation commands.
We argue for an object-centric approach which allows objects to be handled explicitly by the model. Prior work has
encoded objects as bounding box positions~\cite{devin2017deep} for manipulation tasks, but does not use end-to-end training and discards the features of the objects, instead just concatenating their pixel positions. We expand upon this work and evaluate a taxonomy of ``object-centric" neural network models on the driving task.

\section{OBJECT-CENTRIC POLICIES}
We describe a generic architecture that takes in RGB images and outputs actions. Our model expresses a series of choices that provide different object  properties to the model. Our goal is to identify which aspects are important for visuomotor tasks such as autonomous driving. Algorithm~\ref{alg:object} provides pseudo-code for implementing the different variants of our method.

\subsection{Generic Architecture}

The generic form of our model takes in an RGB image and outputs two sets of features: global image contextual features and an object-centric representation.
The global contextual features are produced by a convolutional network over the whole image, followed by a global average pooling operation.
The object-centric representation is constructed as described below to produce a fixed-length object-centric representation.
The global features are concatenated with the object representation, and passed to a fully connected policy network which outputs a discretized action.
For on-policy evaluation, a hard-coded PID controller converts the action to low-level throttle, steer, and brake commands.
\begin{algorithm}
\caption{Discrete Object Centric Representation. \newline
\textbf{sparsity}: bool to discard low-scoring objects.\newline
\textbf{k}: number of objects to keep if sparsity is on.\newline
\textbf{concatenation}: bool to concatenate object features.\newline
\textbf{summation}: bool to sum object features weighted by scores.}
\begin{algorithmic}[1]
  \State image $I$ is received from the sensors
  \State $G$ := GlobalFeatures($I$)
  \State $O$ := Detector($I$) // List of detected bounding boxes
  \For {$O_i \in O$}
  \State $f_i$ := RoI($O_i$) // Object features
  \State $w_i$ := Selector($f_i$, $G$) // Object score
  \EndFor
 \State $\bar{w}_0, ... \bar{w}_N$ = Softmax($w_0, ..., w_N$)
 \State $f_i$ := $\bar{w}_i * f_i$ for all $i$
 \If {sparsity}
   \State Sort objects by $\bar{w}$ and keep only the top $k$
  \EndIf
  \If {concatenation}
  \State \Return concatenate(remaining $f_i$, sorted by $w_i$)
  \ElsIf {summation}
    \State \Return sum(remaining $f_i$)
  \EndIf
\end{algorithmic}
\label{alg:object}
\end{algorithm}

\subsection{Objectness Taxonomy}
\label{sec:taxonomy}
What does it mean for an end-to-end model to be ``object-centric"?
In this section, we define a taxonomy of structures that leverage different aspects of ``objectness".
By defining this taxonomy and placing previous work within it, we evaluate which aspects bring the greatest gains in performance in various driving scenarios.
The aspects discussed are \emph{countability}, \emph{selection}, and \emph{aggregation}. Figure~\ref{fig:taxonomy} visualizes our representation taxonomy.

\subsubsection{Countability: Discrete vs Continuous}
An example of a continuous object-centric representation is a pixel-level attention map over an image, as used in~\cite{kim2017interpretable}.
In contrast, a discrete representation could be a bounding box or instance mask.
The potential benefit of keeping a discrete object structure is that a model may need to reason explicitly over instances (such as cars navigating an intersection) rather than reasoning over a bag-of-vehicles representation.
Our implementation of discrete objects applies a pre-trained FPN detector \cite{lin2017feature} to output bounding boxes for vehicles and pedestrians. We utilize an RoI-pooling layer \cite{girshick2015fast} to extract region features for each box. The boxes and their respective features are treated as a set of objects.
In the discrete setting, we define $O$ as the list of objects returned by the detector, and $f(o_i)$ as the RoI features of the $i$-th object. We define $G$ as the global features from the whole image.

\subsubsection{Selection: Sparse vs Dense}
Should the policy model reason over all objects at once (dense), or should it first select a fixed number (sparse) of salient objects and consider only those?
The former allows more flexibility, but \eg may distract the policy with cars that are very far away or separated from the agent by a median.
To obtain a relevance score for each object, we train a task-specific selector jointly with the policy to minimum the behavioral cloning loss. The selector is a network that takes in the RoI features of each object concatenated with the global image features and outputs a scalar score, indicating the relevance of the object. The scores $w$ are evaluated with a softmax to produce a weight between 0 and 1 for each object.
In the sparse model, only the top $k$ scoring objects are used in the policy.

\subsubsection{Aggregation: Sum vs Concatenate}
If using discrete objects, a decision needs to be taken about how to combine the objects into a single representation.
One possible approach is to weight and sum the features of the objects, while another approach is to concatenate the features.
The former is agnostic to the number of objects and is order invariant, while the latter may allow for more nuanced computation about multi-object decisions. Our implementation of the concatenation approach is to sort the objects by their selector weights and concatenate the features $\bar{w}_i * f_i$ in order from largest $w_i$ to smallest.

\begin{figure*}[ht]
\begin{center}
\includegraphics[width=\textwidth]{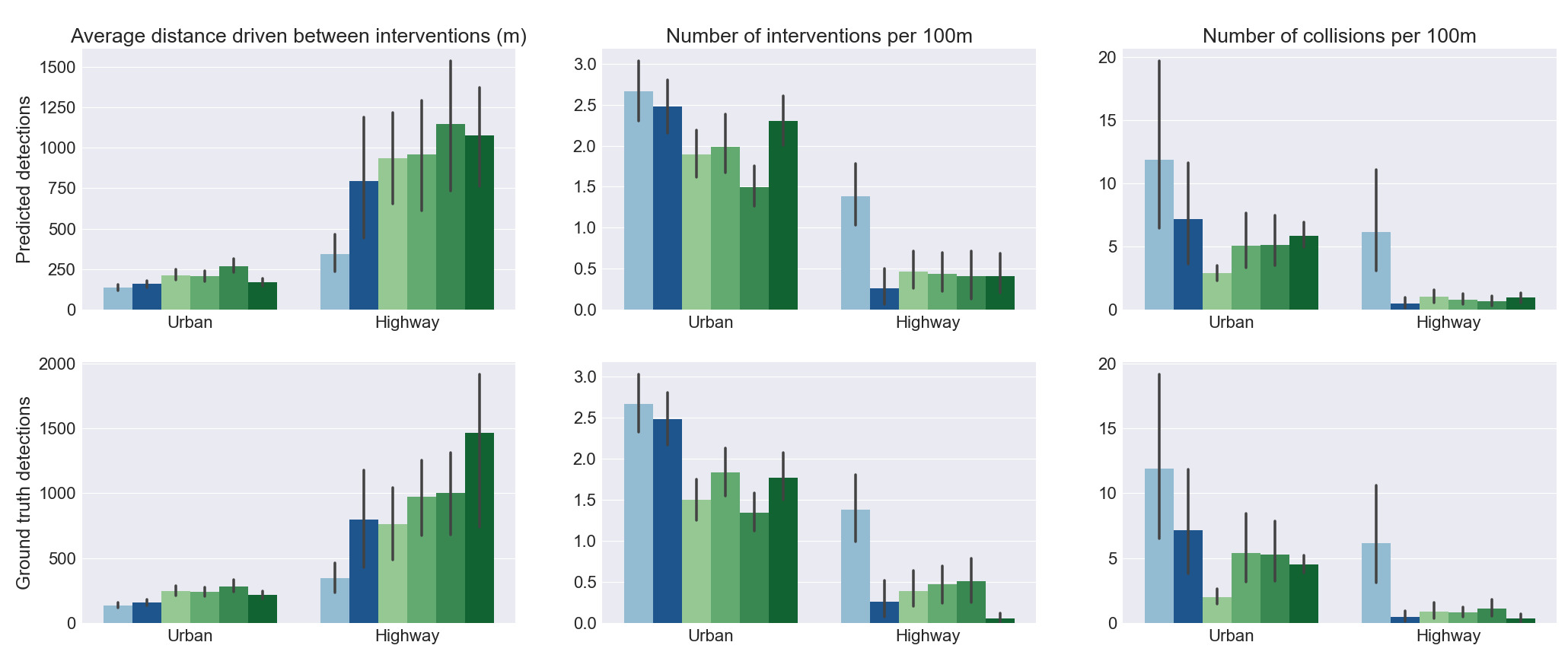}
\includegraphics[width=\textwidth]{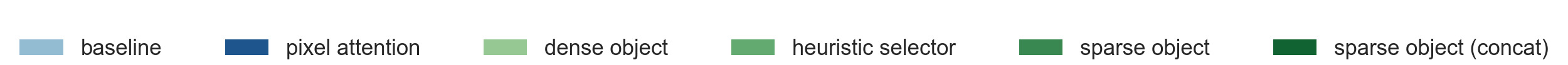}
\end{center}
\vspace{-3mm}
\caption{Driving performance. From left to right: driving distance between interventions, number of interventions per 100m, number of collisions per 100m. The top row shows results using a learned detection model, while the bottom row uses ground-truth bounding box. The object centric models (green) overall perform better than the object agnostic models (blue), with the sparse models being the best. The highway environment is easier to drive than the urban environment. Comparing the heuristic selector with the learned selector used in the ``sparse object" model, it is clear that learning a selector provides better results. }
\label{fig:onpolicy}
\end{figure*}

\section{EXPERIMENTS}

We evaluate our object-centric models on both a simulated environment and a real-world dataset. Specifically, we use the Grand Theft Auto V simulation \cite{krahenbuhl2018free} and the Berkeley DeepDrive Video dataset \cite{xu2017end} for online and offline evaluation, respectively. All models are
trained on a behavioral cloning objective.

\subsection{Evaluation Setup}
\subsubsection{Online Driving Simulation}
For the simulation experiments, $1.6$ million training frames were collected by using the in-game
navigation system as the expert policy. Following a DAgger-like~\cite{ross2011reduction} augmented imitation learning pipeline, noise was added to the control command every 30 seconds to generate diverse behavior. The noisy control frames and the following $\sim 7$ frames were dropped during training to avoid replicating noisy behavior. The simulation was rendered at 12 frames per second. The training dataset was collected over 1000 random paths across 2km in the game. The in-game times ranged from 8:00 am to 7:00 pm with the default weather condition set to ``cloudy".

In total,
Each frame included control signals, such as speed, angle, throttle, steering, brake, as well as ground-truth bounding boxes around vehicles and pedestrians. During our training and testing procedure we used a camera in front of the car which keeps a fixed $60^\circ$ horizontal field of view (FoV). The maximum speed of all vehicles was set to 20km/h.

When training a policy, the expert's continuous action was discretized into 9 actions: (\textit{left, straight, right}) $\times$ (\textit{fast, slow, stop}). At evaluation time, we used a PID controller (shared for all models) to translate the discrete actions into continuous control signals per frame.

\begin{figure*}[ht]
\begin{center}
        \includegraphics[width=0.233\linewidth]{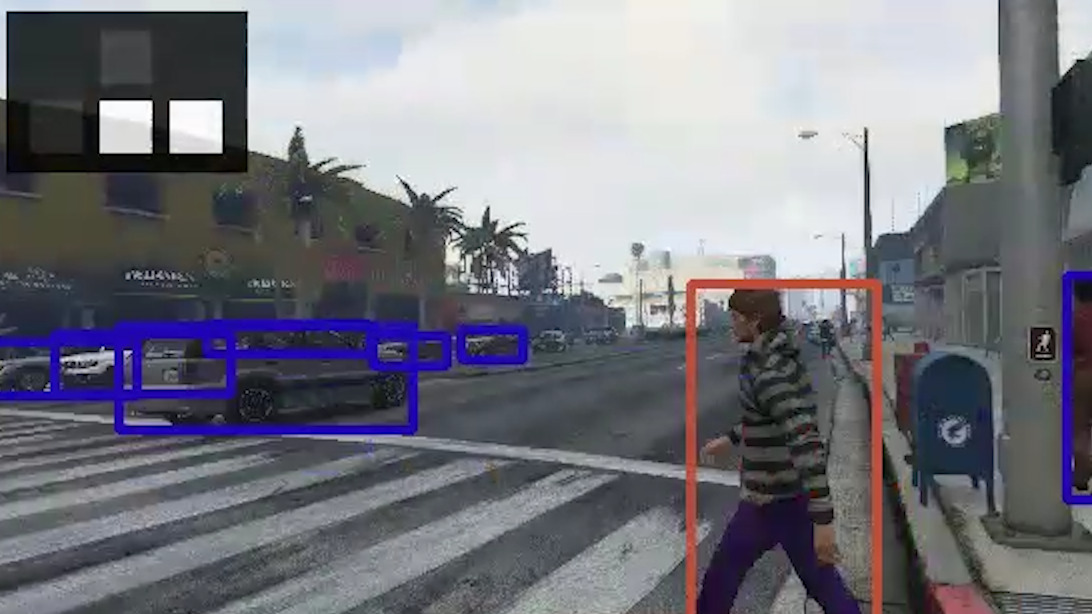}
    \includegraphics[width=0.233\linewidth]{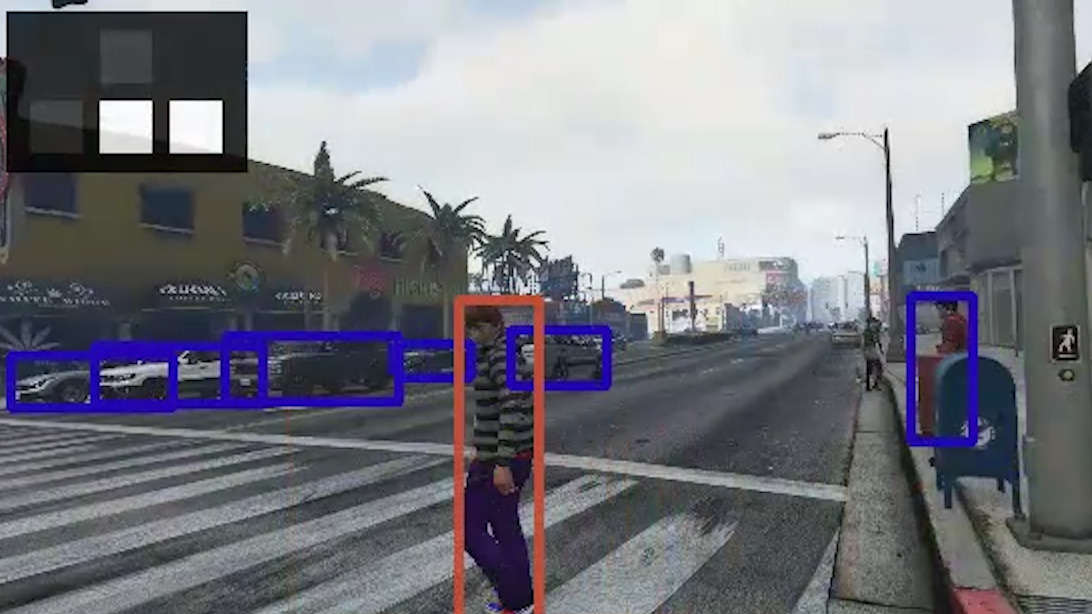}
    \includegraphics[width=0.233\linewidth]{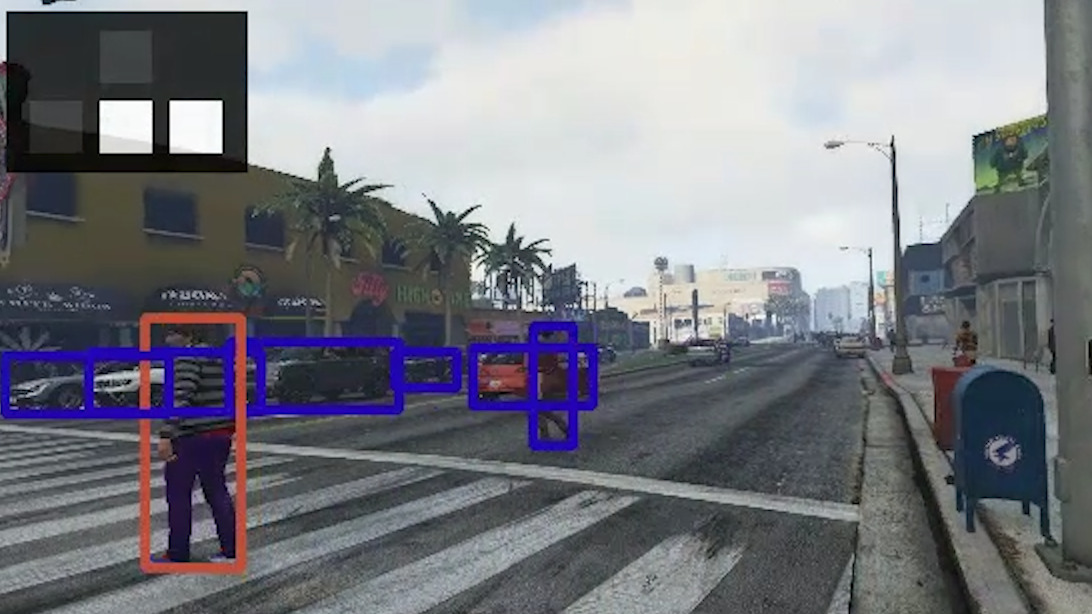}
    \includegraphics[width=0.233\linewidth]{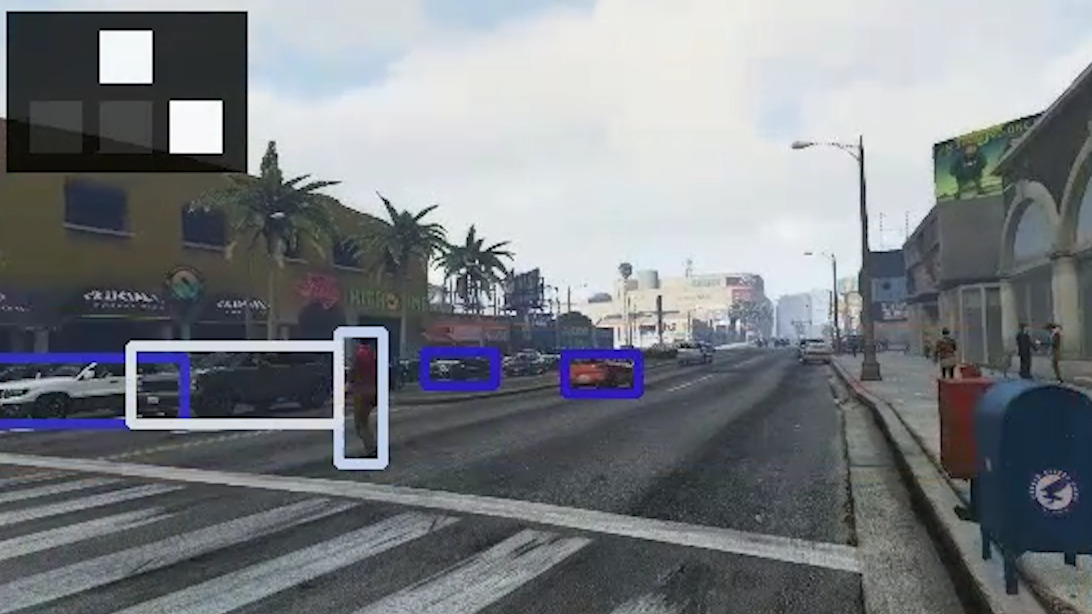}

    \includegraphics[width=0.233\linewidth]{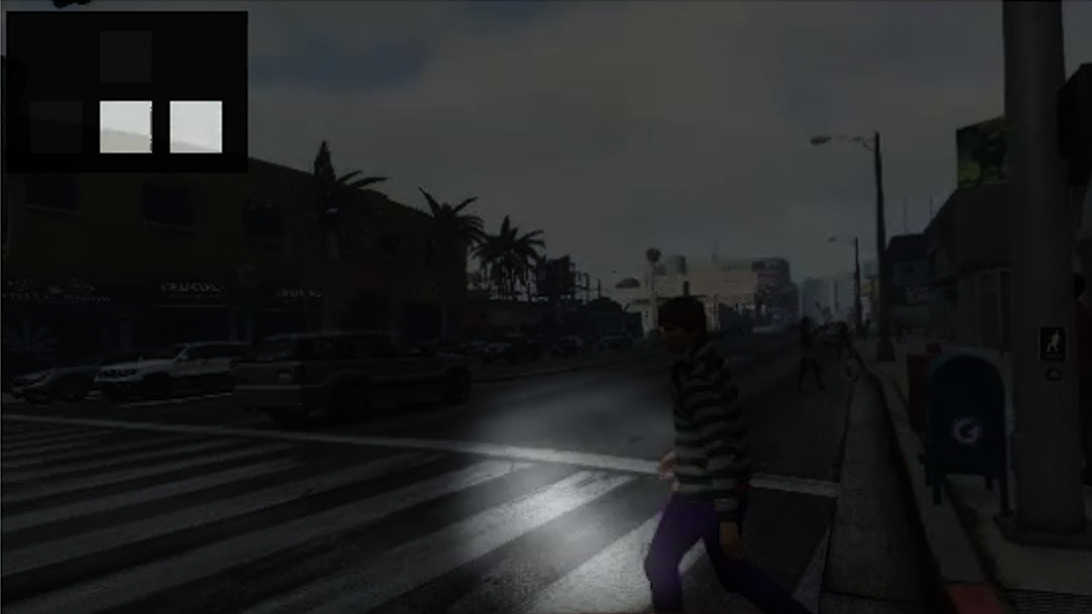}
    \includegraphics[width=0.233\linewidth]{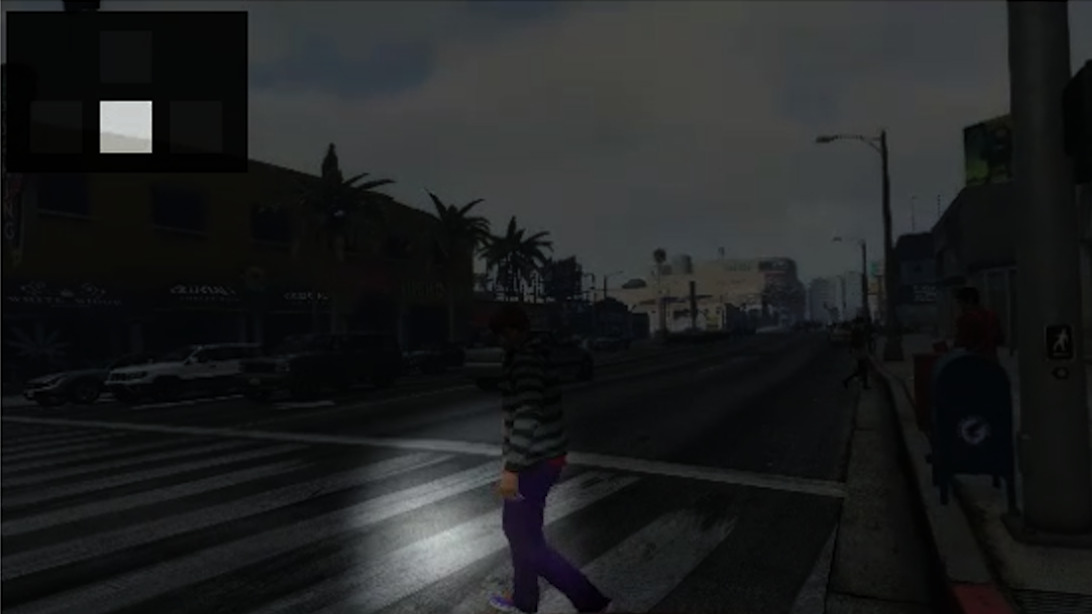}
    \includegraphics[width=0.233\linewidth]{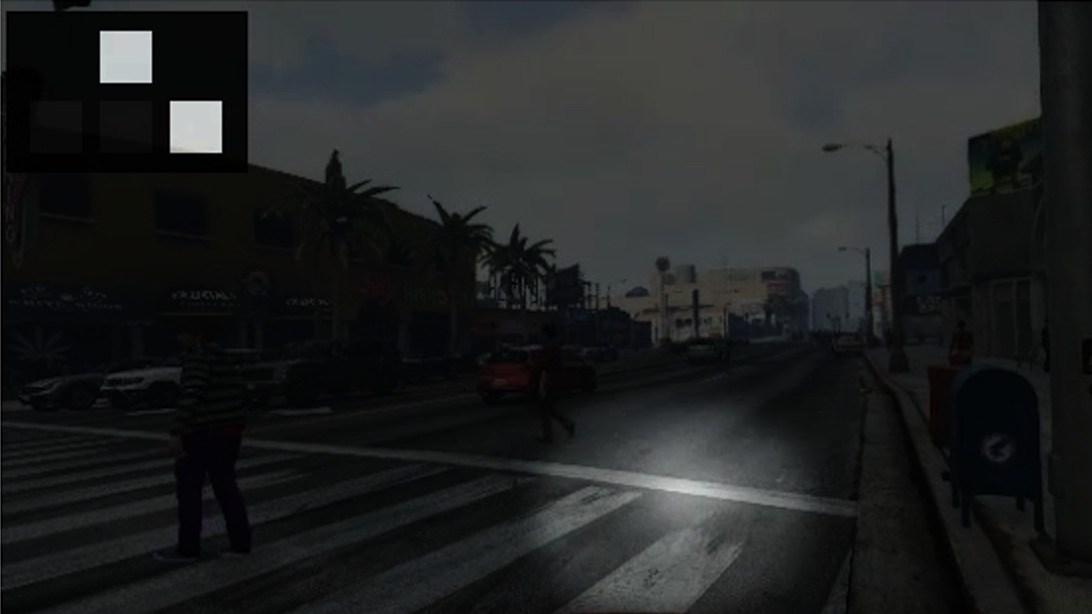}
    \includegraphics[width=0.233\linewidth]{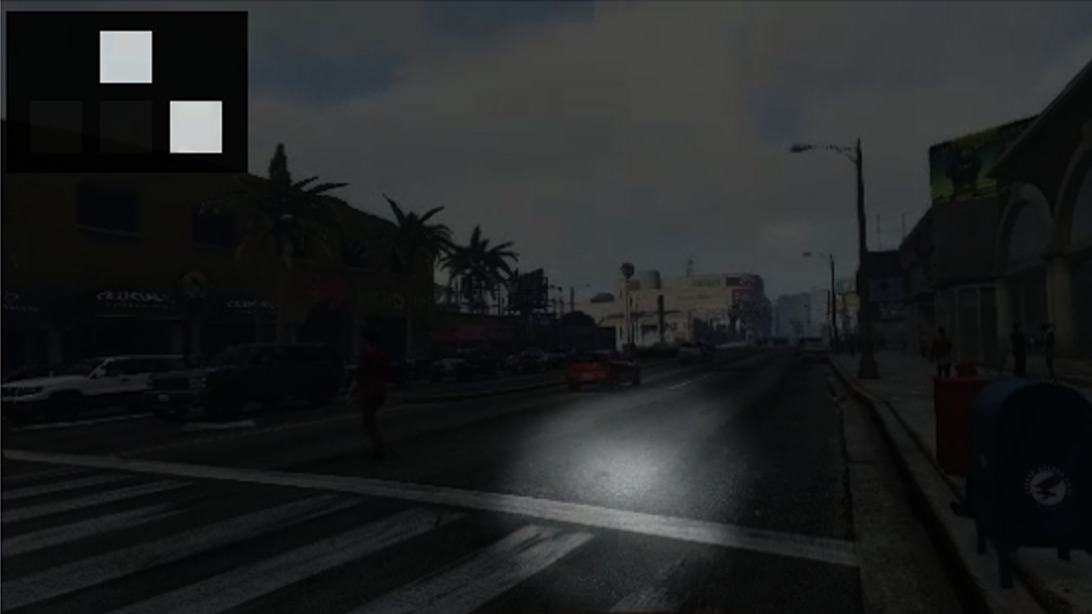}

    \includegraphics[width=0.233\linewidth]{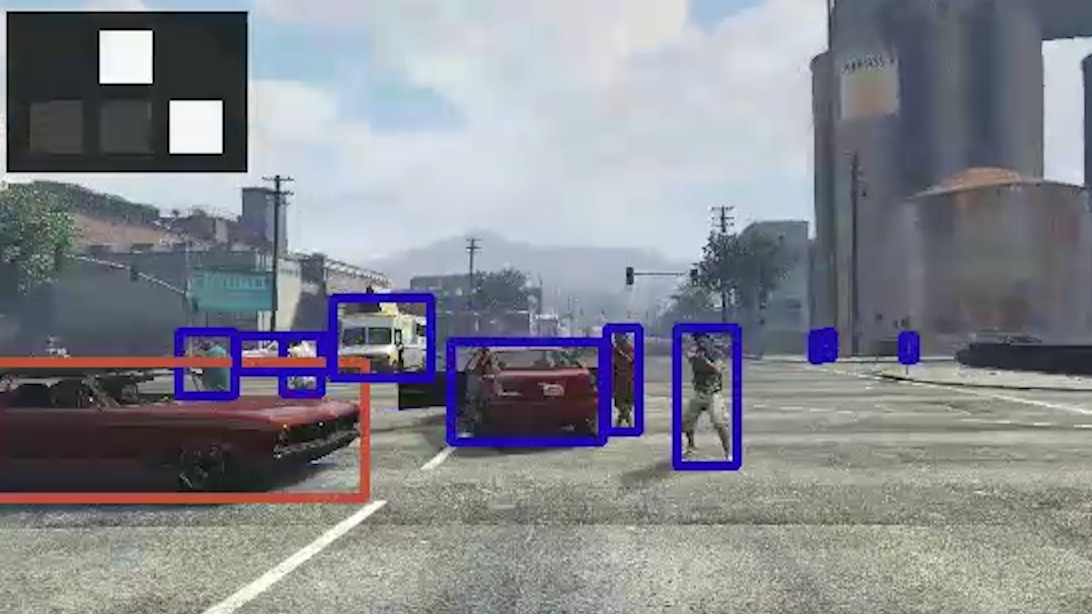}
    \includegraphics[width=0.233\linewidth]{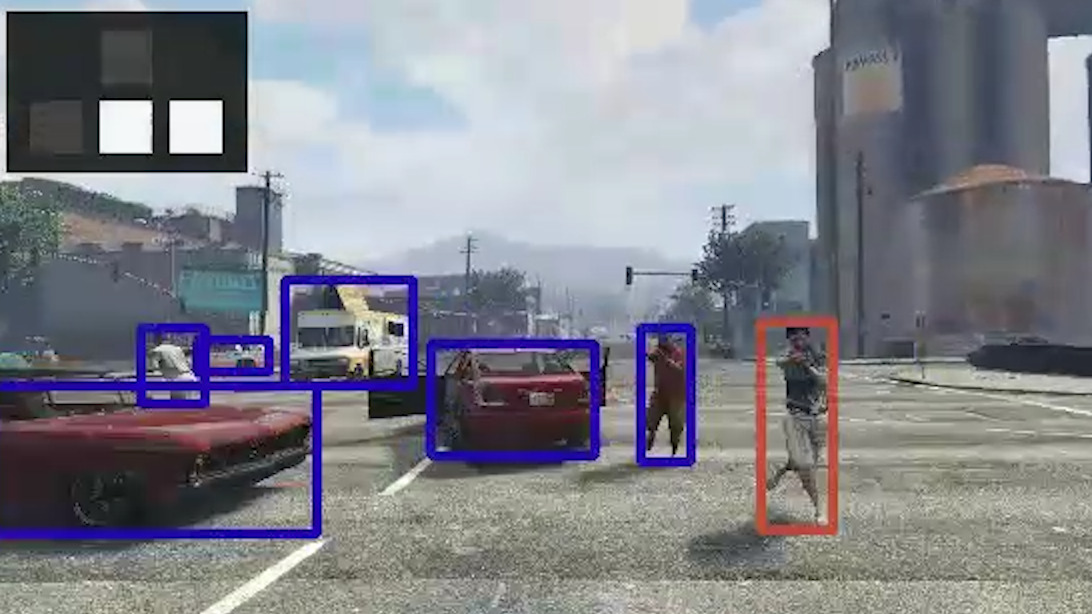}
    \includegraphics[width=0.233\linewidth]{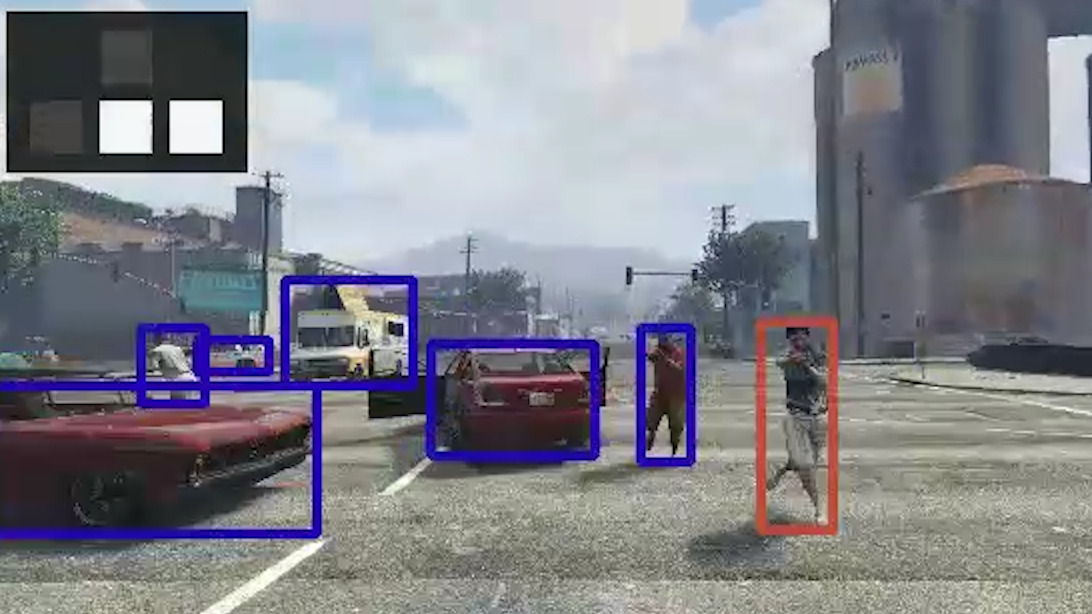}
    \includegraphics[width=0.233\linewidth]{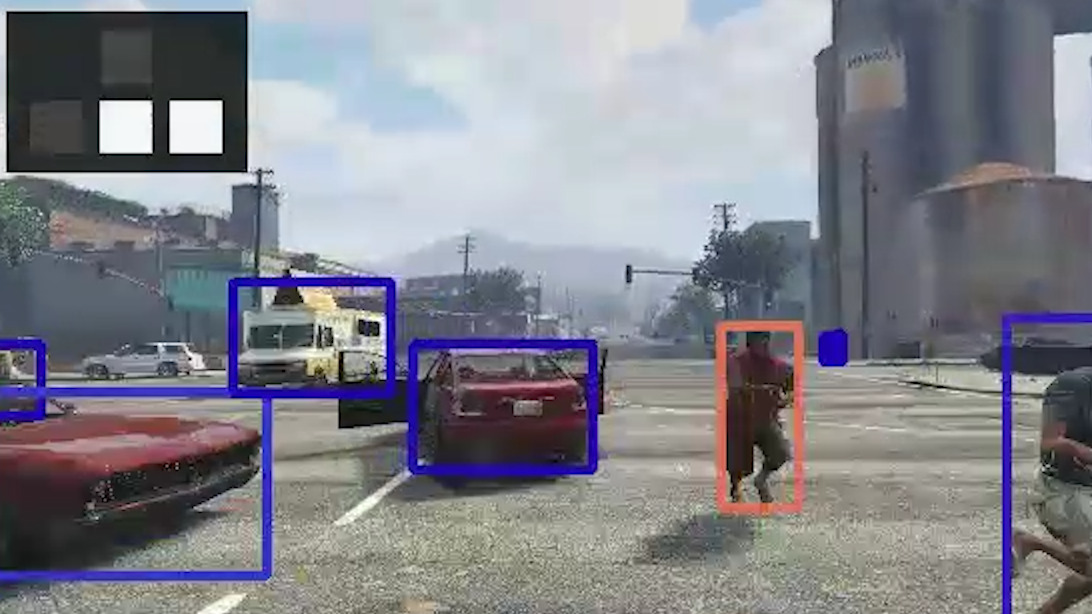}

    \includegraphics[width=0.233\linewidth]{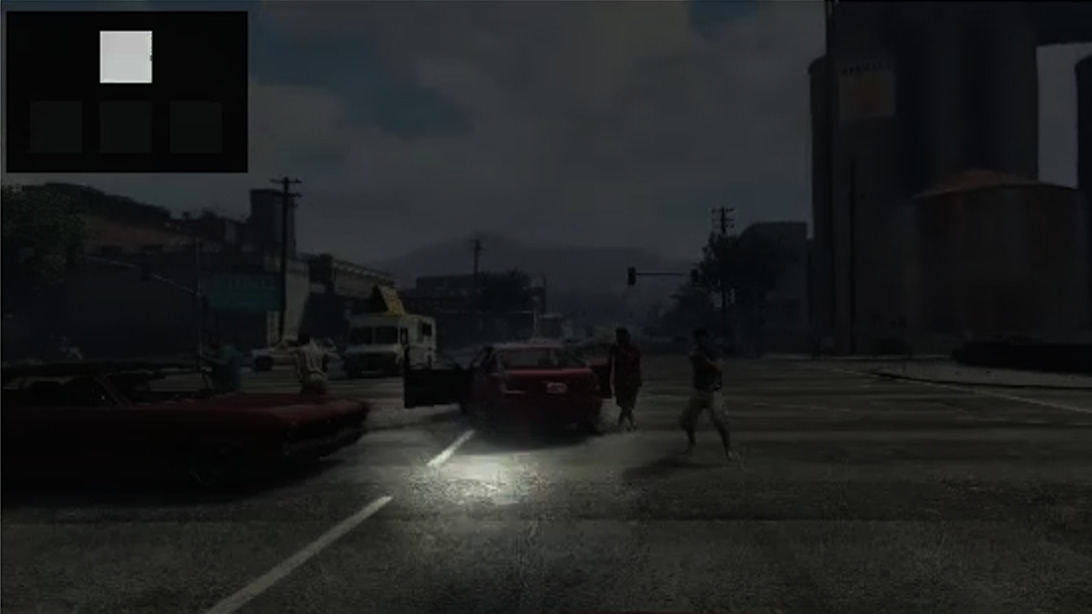}
    \includegraphics[width=0.233\linewidth]{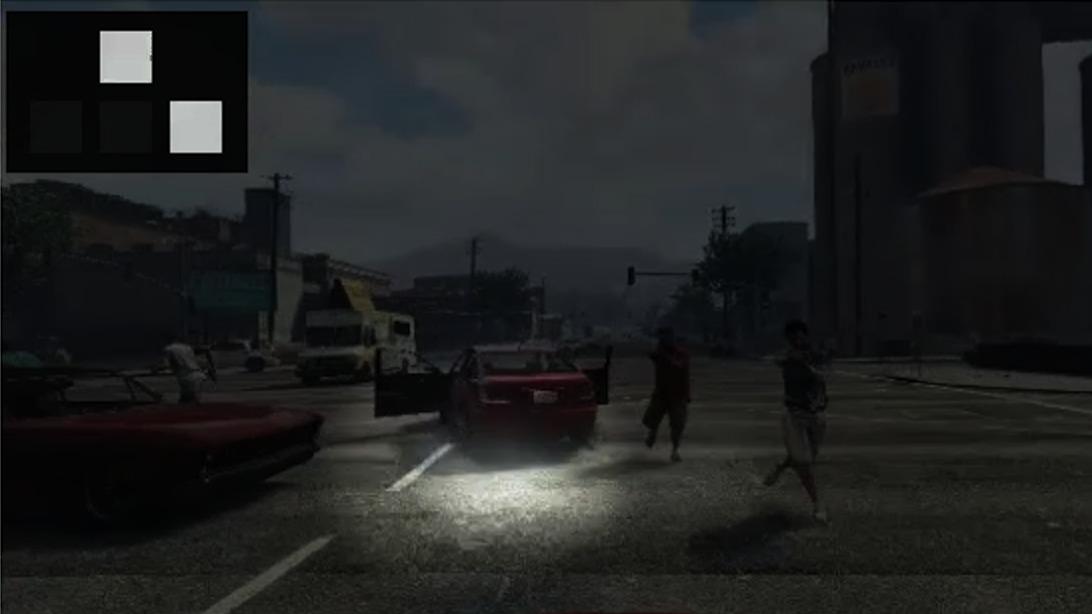}
    \includegraphics[width=0.233\linewidth]{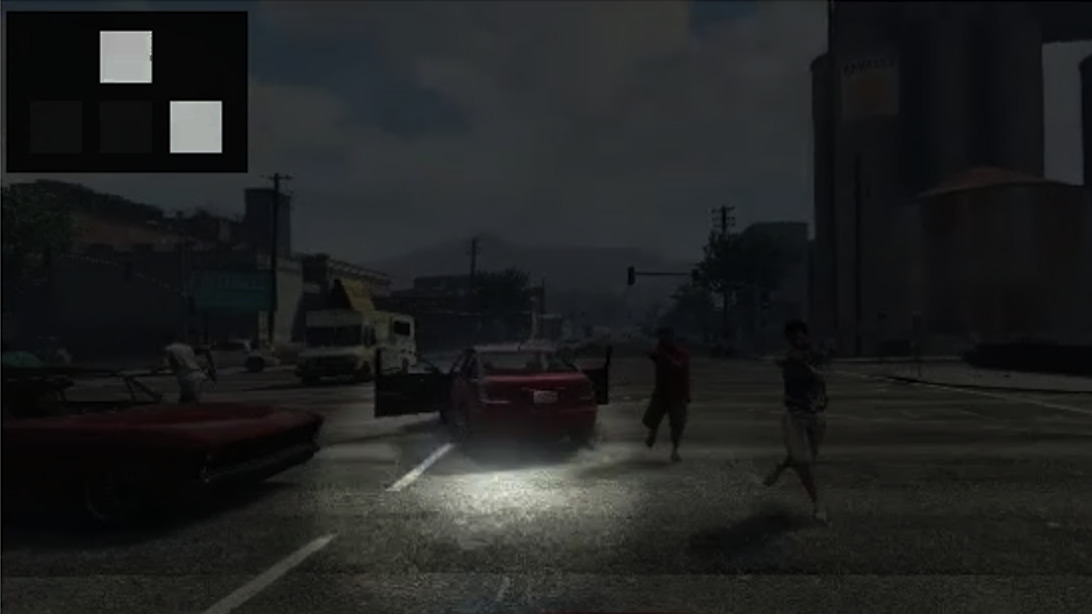}
    \includegraphics[width=0.233\linewidth]{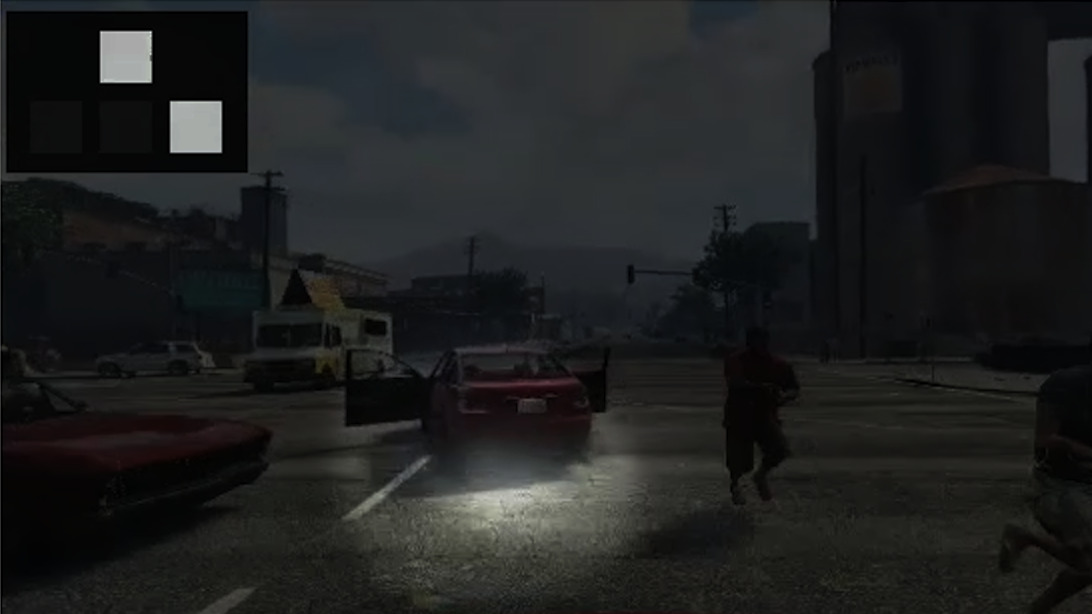}
    \end{center}
                \caption{Sample scenes from the Grand Theft Auto V simulation with our sparse model's learned object selector compared against a learned pixel-level attention. For rows 1 and 3, red indicates a high scoring object, and blue is low scoring (best viewed on screen). For rows 2 and 4,
                then pixel attention is shown by brightness of the pixels.
                The actions output by each model are shown by the white squares in the corners: accelerator is the top square, and the bottom squares are turn left, brake, and turn right, respectively. A single action may both turn and accelerate or brake.
                Rows 1 and 2 shows both models performing well, while rows 3 and 4 show the pixel attention model ignoring pedestrians and deciding to accelerate towards them. The object centric model is more conservative and attends strongly to the pedestrians, choosing to slow down instead of speeding up.}
    \label{fig:selection}
\end{figure*}

For testing, we deployed the model in $8$ locations unseen during training: $2$ highway and $6$ urban intersections.
Figure~\ref{fig:selection} demonstrates some example scene layouts in our simulation environment.
For each location, we tested the model for $100$ minutes: the agent was run for 10 independent roll-outs lasting 10 minutes each.
If the vehicle crashed or got stuck during a rollout, the incident was recorded and the in-game AI intervened over for at least 15 seconds until it recovered.
An extreme accident which took more time to recover from would be penalized more in our metric as it would travel less overall distance; the frames during the intervention were not counted towards the total.

The models were evaluated with several metrics.
For each roll-out, we calculated the total distance travelled, the number of collisions, and the number of interventions by the in-game AI.
To compare across roll-outs, we computed the distance driven between AI interventions, the number of collisions and interventions per 100m traveled.
\begin{figure*}[ht]
\begin{center}
\includegraphics[width=0.325\linewidth]{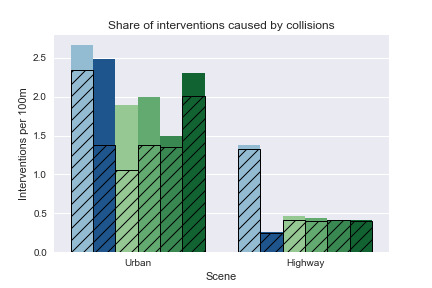}
\includegraphics[width=0.325\linewidth]{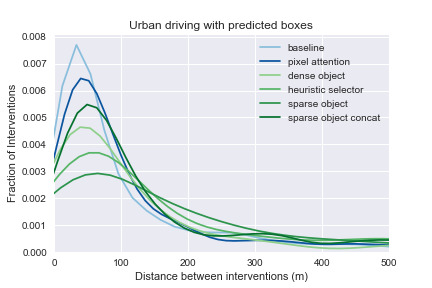}
\includegraphics[width=0.325\linewidth]{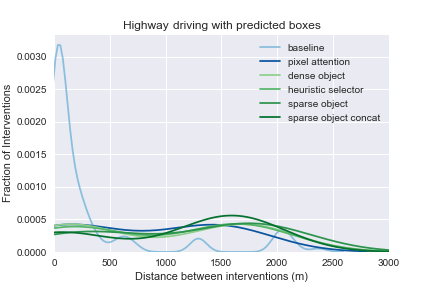}
\end{center}
\vspace{-3.5mm}
    \caption{Analysis of intervention frequency. Same legend as Figure~\ref{fig:onpolicy}. On the left, the shaded region indicates the proportion of interventions caused by collisions. In the highway environment, almost all interventions are cause by collisions, but in the urban environment, the policy gets stuck at intersections, as shown in the supplementary video.
    On the right, histograms shows how far each model drove between interventions and collisions. In the urban environment the object centric approaches drove farthest before an intervention. In the highway environment, the pixel attention performs slightly better.}
    \label{fig:urban}
\end{figure*}
\begin{figure*}[ht]
    \begin{center}
    \includegraphics[width=\linewidth]{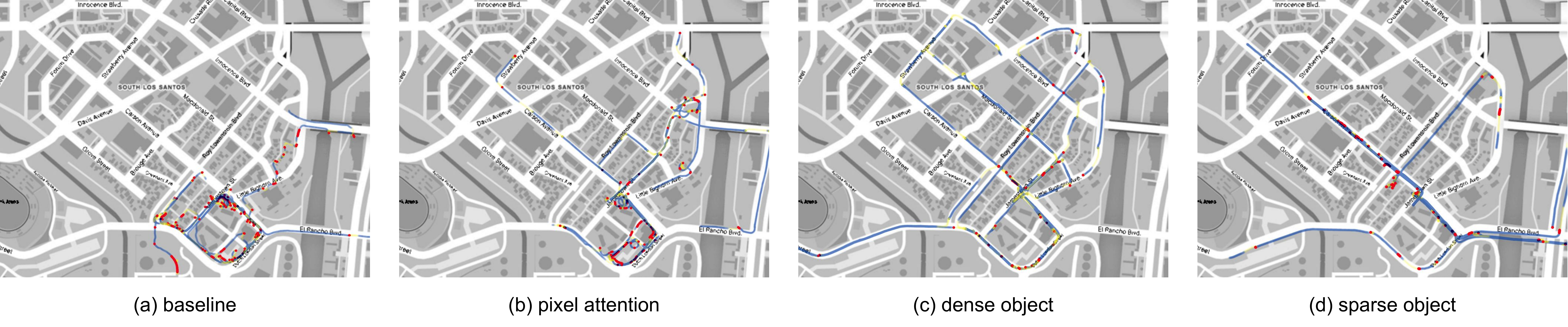}

\end{center}
\vspace{-4mm}
    \caption{Sample trajectories from the evaluation. Yellow dots indicate to interventions while red dots indicate collisions (best viewed on screen). This example illustrates the reliability of the object centric models over the the baselines, with fewer collisions \& interventions and greater road coverage.}
    \label{fig:GPS}
\end{figure*}

\subsubsection{Real-world Offline Dataset}

We used $2.2$ million training frames and $0.2$ million testing frames from a large-scale crowd-sourcing dash-cam video dataset, with diverse driving behaviors.
Each frame was accompanied by raw sensory data from GPS, IMU, gyroscope, magnetometer, as well as sensor-fused measurements like course and speed.

As in Xu et al~\cite{xu2017end}, the model was trained to predict the expert's future linear and angular speeds for each frame at intervals of $1/3$ seconds during training. For evaluation,  speed and angle were discretized into $30$ bins each and then were mapped into the joint distribution of speed and angle into $30\times30=900$ bins.
Following the method in Xu et al, we evaluated the $900$-way classification model by the perplexity of the model on withheld test data.
Specifically, we calculated the value of softmax loss function as perplexity indicator.

\begin{table}[h]
\ra{1.2}
\begin{center}
\begin{tabular}{|l|c|c|c|c|c|}
\hline
\% data trained on & 5\% & 10\% & 25\% & 50\% & 100\% \\
\hline
baseline & 2.52 & 2.40 & 2.29 & 1.94 & \textbf{1.80} \\ 
pixel attention & 2.70 & 2.33 & 2.15 & 1.96 & 1.84 \\ 
dense object & 2.34 & 2.24 & 2.07 & 2.06 & 2.01 \\ 
heuristic selector & 2.48 & 2.39 & 2.31 & 2.13 & 2.10 \\ 
sparse object & \textbf{2.31} & \textbf{2.23} & 2.19 & 2.07 & 2.10 \\ 
sparse object concat & 2.37 & 2.31 & \textbf{2.04} & \textbf{1.93} & 1.82 \\ \hline
\end{tabular}

\end{center}
\caption{Sparse training real world evaluation. To evaluate the models trained on real images, 
we measure the perplexity of the models on withheld test data as an off-policy evaluation. Lower perplexity 
indicates that dataset was modeled more accurately.
\label{table:realperplexity}}
\end{table}

\subsection{Implementation Details}
The convolutional backbone is a 34-layer DLA model~\cite{yu2018deep} pre-trained on ImageNet~\cite{russakovsky2015imagenet}, with the open-source framework PyTorch~\cite{paszke2017automatic}. For all models, including the baseline and attention models, this network is trained end-to-end as part of the policy.
We use a Detectron model~\cite{Detectron2018} trained on MSCOCO~\cite{lin2014microsoft} to generate bounding boxes for moving objects, specifically vehicles and pedestrians.
We used the Adam optimizer~\cite{kingma2014adam} for $3$ epochs with initial learning rate $0.001$,  weight decay $10^{-4}$, and batch size 128.
Unlike in ~\cite{codevilla2017end, sauer2018conditional}, we do not use any data augmentation. All sparse models use $k=5$ to keep the top $5$ objects and discard the rest. The selector is a linear mapping from object features to object score, optimized jointly with the features and the policy.
A final linear layer maps from the  concatenation of the object and global features to the predicted action.

\subsection{Results}
We evaluate several baselines, prior methods, and ablations.
The \emph{baseline} method does not represent objects or use attention at inference time.
The \emph{pixel attention} method is the same as \emph{baseline} but with an additional pixel-level attention mechanism, learned end-to-end with the task.

Next, we evaluate several object-centric models drawn from our taxonomy. The results labeled \emph{dense object} use a discrete and dense object representation with summation of the objects weighted by a learned selector. \emph{Sparse object} is the same as \emph{dense object}, but only looks at the top 5 objects in the scene, as scored by the learned selector. While the preceding models used the selector to weight object features before summing them, \emph{sparse object concat} concatenates the features of the top 5 objects and passes the entire list to the fully connected policy.
We also evaluate our selector by comparing to a heuristic selector: the size of the object's bounding box. The results using the heuristic selector in a sparse object model are labeled \emph{heuristic selector}.

The results of the on-policy simulated driving are shown in Figure~\ref{fig:onpolicy}. We show several metrics: the number of collisions, the number of times the agent got stuck, and the distance driven between these. Each evaluation was repeated for two environments: urban (which has many intersections and cars/pedestrians) and highway (which is mostly driving straight). The object-centric methods consistently outperform the two object-agnostic method in the urban evaluation, while the highway environment shows good performance for all attention models.

The comparable performance between the evaluation with ground truth boxes versus predicted boxes (from a detector trained on MSCOCO~\cite{lin2014microsoft}) indicates that our method is robust to noisy detections. Figure~\ref{fig:GPS} visualizes evaluation roll-outs along a map with collisions and interventions drawn in. These maps show how the object centric models drive for longer without crashing or getting stuck, and how they end up farther from their start point than the baseline and pixel attention models. This is supported by the histograms of distance between interventions in Figure~\ref{fig:urban} which shows how the sparse models especially drive farther between interventions.

\begin{figure*}[ht]
    \begin{center}
    \ra{1.3}

\includegraphics[width=0.233\linewidth]{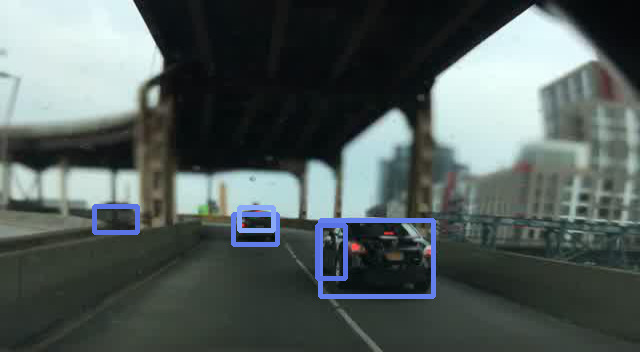}
\includegraphics[width=0.233\linewidth]{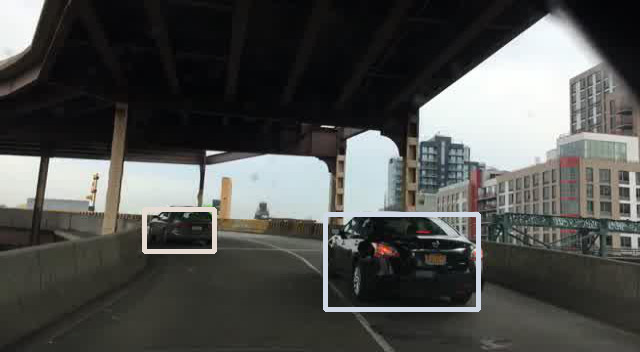}
\includegraphics[width=0.233\linewidth]{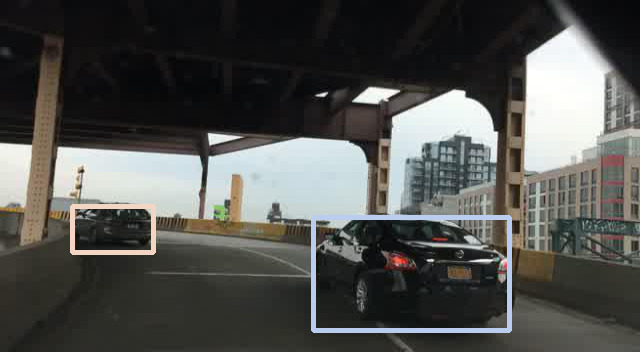}
\includegraphics[width=0.233\linewidth]{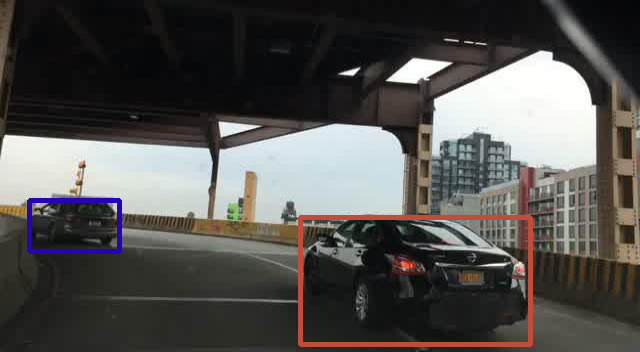}

\end{center}
    \caption{Sample scenes from the Berkeley DeepDrive Video dataset with the sparse model's learned selector visualized. Red indicates a high scoring object, and blue is low scoring (best viewed on screen). Our method is robust to imperfect detections, such as overlapping bounding boxes, for both both day and night scenes.}
    \label{fig:selectionnexar}
\end{figure*}

To identify the benefits of using a learned selector over boxes, we compared the \emph{sparse object} model against a heuristic selector, which assigns importance to objects based on their size. The motivation for this heuristic is that larger objects are likely to be closer, and therefore more important for the policy. Figure~\ref{fig:onpolicy} shows that the model with a learned selector performs equally or better than the heuristic for every metric. Although some other heuristic may work better, we conclude that learning the selector jointly with the policy is beneficial.

The final experiment in Table~\ref{table:realperplexity} is an off-policy evaluation on the real world dataset that measures the perplexity of the learned model with respect to test data. As we cannot evaluate the policies in the real world, the perplexity provides a proxy for how well the data can be modeled.
When trained on only a subset of the data (from 5\% to 50\%), the sparse object models performs best, with concatenation overtaking summation in the medium data regime.
The concatenation model performs equally well to the baseline once all the data has been seen, indicating that the sparse model is advantageous for low data problems, and that the sparse concat model is ideal for medium to large data situations.
The object prior that our models leverage allows them to learn quickly from little data without being distracted by irrelevant pixels.
Figure~\ref{fig:selectionnexar} shows example scenes with our model's attention.

\section{CONCLUSION}
We defined a taxonomy over object-centric models and showed in an on-policy evaluation that sparse object models outperformed object-agnostic models according to our metrics of distance driven and frequency of collisions and interventions. Our results show that highway driving is significantly easier than navigating intersections; the necessity of navigating city environments showcase the advantages of representing objects. Discreteness and sparsity, along with a learned selection mechanism, seem to be the most important aspects of object-centric models.

For simplicity, this work only considered the presence of vehicles and pedestrians and did not evaluate the policy's ability to  follow the rules of the road. Using generic object detection rather than class specific detection is an obvious direction of future work, and would hopefully lead to paying attention to streetlights, signage, and other objects relevant to driving. These types of objects are crucial for following the rules of the road, and we expect that object-centric policies which attend to these objects would provide even more gains. Promising avenues for future work also include leveraging the 3D nature of objects and their temporal coherence.

\newpage
\section{ACKNOWLEDGEMENTS}
This work was supported by Berkeley AI Research and Berkeley Deep Drive. Coline Devin is supported by an NSF Graduate Research Fellowship.

\addtolength{\textheight}{-12cm}   

\bibliographystyle{IEEEtran}
\bibliography{ref.bib}

\begin{thebibliography}{10}
\providecommand{\url}[1]{#1}
\csname url@rmstyle\endcsname
\providecommand{\newblock}{\relax}
\providecommand{\bibinfo}[2]{#2}
\providecommand\BIBentrySTDinterwordspacing{\spaceskip=0pt\relax}
\providecommand\BIBentryALTinterwordstretchfactor{4}
\providecommand\BIBentryALTinterwordspacing{\spaceskip=\fontdimen2\font plus
\BIBentryALTinterwordstretchfactor\fontdimen3\font minus
  \fontdimen4\font\relax}
\providecommand\BIBforeignlanguage[2]{{%
\expandafter\ifx\csname l@#1\endcsname\relax
\typeout{** WARNING: IEEEtran.bst: No hyphenation pattern has been}%
\typeout{** loaded for the language `#1'. Using the pattern for}%
\typeout{** the default language instead.}%
\else
\language=\csname l@#1\endcsname
\fi
#2}}

\bibitem{muller2018driving}
M.~M{\"u}ller, A.~Dosovitskiy, B.~Ghanem, and V.~Koltun, ``Driving policy
  transfer via modularity and abstraction,'' \emph{arXiv preprint
  arXiv:1804.09364}, 2018.

\bibitem{bojarski2016end}
M.~Bojarski, D.~Del~Testa, D.~Dworakowski, B.~Firner, B.~Flepp, P.~Goyal, L.~D.
  Jackel, M.~Monfort, U.~Muller, J.~Zhang, \emph{et~al.}, ``End to end learning
  for self-driving cars,'' \emph{arXiv preprint arXiv:1604.07316}, 2016.

\bibitem{bojarski2017explaining}
M.~Bojarski, P.~Yeres, A.~Choromanska, K.~Choromanski, B.~Firner, L.~Jackel,
  and U.~Muller, ``Explaining how a deep neural network trained with end-to-end
  learning steers a car,'' \emph{arXiv preprint arXiv:1704.07911}, 2017.

\bibitem{xu2017end}
H.~Xu, Y.~Gao, F.~Yu, and T.~Darrell, ``End-to-end learning of driving models
  from large-scale video datasets,'' in \emph{CVPR}, 2017.

\bibitem{huval2015empirical}
B.~Huval, T.~Wang, S.~Tandon, J.~Kiske, W.~Song, J.~Pazhayampallil,
  M.~Andriluka, P.~Rajpurkar, T.~Migimatsu, R.~Cheng-Yue, \emph{et~al.}, ``An
  empirical evaluation of deep learning on highway driving,'' \emph{arXiv
  preprint arXiv:1504.01716}, 2015.

\bibitem{thrun2006stanley}
S.~Thrun, M.~Montemerlo, H.~Dahlkamp, D.~Stavens, A.~Aron, J.~Diebel, P.~Fong,
  J.~Gale, M.~Halpenny, G.~Hoffmann, \emph{et~al.}, ``Stanley: The robot that
  won the darpa grand challenge,'' \emph{Journal of Field Robotics}, 2006.

\bibitem{urmson2008autonomous}
C.~Urmson, J.~Anhalt, D.~Bagnell, C.~Baker, R.~Bittner, M.~Clark, J.~Dolan,
  D.~Duggins, T.~Galatali, C.~Geyer, \emph{et~al.}, ``Autonomous driving in
  urban environments: Boss and the urban challenge,'' \emph{Journal of Field
  Robotics}, 2008.

\bibitem{ziegler2014making}
J.~Ziegler, P.~Bender, M.~Schreiber, H.~Lategahn, T.~Strauss, C.~Stiller,
  T.~Dang, U.~Franke, N.~Appenrodt, C.~G. Keller, \emph{et~al.}, ``Making
  bertha drive-an autonomous journey on a historic route.'' \emph{ITSM}, 2014.

\bibitem{pomerleau1989alvinn}
D.~A. Pomerleau, ``Alvinn: An autonomous land vehicle in a neural network,'' in
  \emph{NIPS}, 1989.

\bibitem{muller2006off}
U.~Muller, J.~Ben, E.~Cosatto, B.~Flepp, and Y.~L. Cun, ``Off-road obstacle
  avoidance through end-to-end learning,'' in \emph{NIPS}, 2006.

\bibitem{codevilla2017end}
F.~Codevilla, M.~M{\"u}ller, A.~Dosovitskiy, A.~L{\'o}pez, and V.~Koltun,
  ``End-to-end driving via conditional imitation learning,'' \emph{arXiv
  preprint arXiv:1710.02410}, 2017.

\bibitem{chen2015deepdriving}
C.~Chen, A.~Seff, A.~Kornhauser, and J.~Xiao, ``Deepdriving: Learning
  affordance for direct perception in autonomous driving,'' in \emph{ICCV},
  2015.

\bibitem{sauer2018conditional}
A.~Sauer, N.~Savinov, and A.~Geiger, ``Conditional affordance learning for
  driving in urban environments,'' \emph{arXiv preprint arXiv:1806.06498},
  2018.

\bibitem{devin2017deep}
C.~Devin, P.~Abbeel, T.~Darrell, and S.~Levine, ``Deep object-centric
  representations for generalizable robot learning,'' in \emph{ICRA}, 2017.

\bibitem{kim2017interpretable}
J.~Kim and J.~Canny, ``Interpretable learning for self-driving cars by
  visualizing causal attention,'' in \emph{ICCV}, 2017.

\bibitem{lin2017feature}
T.-Y. Lin, P.~Doll{\'a}r, R.~B. Girshick, K.~He, B.~Hariharan, and S.~J.
  Belongie, ``Feature pyramid networks for object detection.'' in \emph{CVPR},
  2017.

\bibitem{girshick2015fast}
R.~Girshick, ``Fast r-cnn,'' in \emph{ICCV}, 2015.

\bibitem{krahenbuhl2018free}
P.~Kr{\"a}henb{\"u}hl, ``Free supervision from video games,'' in \emph{CVPR},
  2018.

\bibitem{ross2011reduction}
S.~Ross, G.~Gordon, and D.~Bagnell, ``A reduction of imitation learning and
  structured prediction to no-regret online learning,'' in \emph{AISTATS},
  2011.

\bibitem{yu2018deep}
F.~Yu, D.~Wang, E.~Shelhamer, and T.~Darrell, ``Deep layer aggregation,'' in
  \emph{CVPR}, 2018.

\bibitem{russakovsky2015imagenet}
O.~Russakovsky, J.~Deng, H.~Su, J.~Krause, S.~Satheesh, S.~Ma, Z.~Huang,
  A.~Karpathy, A.~Khosla, M.~Bernstein, \emph{et~al.}, ``Imagenet large scale
  visual recognition challenge,'' \emph{IJCV}, 2015.

\bibitem{paszke2017automatic}
A.~Paszke, S.~Gross, S.~Chintala, G.~Chanan, E.~Yang, Z.~DeVito, Z.~Lin,
  A.~Desmaison, L.~Antiga, and A.~Lerer, ``Automatic differentiation in
  pytorch,'' in \emph{NIPS-W}, 2017.

\bibitem{Detectron2018}
R.~Girshick, I.~Radosavovic, G.~Gkioxari, P.~Doll\'{a}r, and K.~He,
  ``Detectron,'' \url{https://github.com/facebookresearch/detectron}, 2018.

\bibitem{lin2014microsoft}
T.-Y. Lin, M.~Maire, S.~Belongie, J.~Hays, P.~Perona, D.~Ramanan,
  P.~Doll{\'a}r, and C.~L. Zitnick, ``Microsoft coco: Common objects in
  context,'' in \emph{ECCV}, 2014.

\bibitem{kingma2014adam}
D.~P. Kingma and J.~Ba, ``Adam: A method for stochastic optimization,''
  \emph{arXiv preprint arXiv:1412.6980}, 2014.

\end{thebibliography}

\end{document}